\documentclass[11pt,letterpaper]{article}
\usepackage{naaclhlt2016}
\usepackage{pslatex}
\usepackage{latexsym}
\usepackage{xspace}
\usepackage{multirow}
\usepackage{amsfonts}
\usepackage{url}
\usepackage{epstopdf}
\usepackage{amsmath}
\usepackage{caption}
\usepackage{tabularx}
\usepackage{pbox}
\usepackage{color}
\usepackage{xcolor}
\usepackage{url}

\naaclfinalcopy 
\def\naaclpaperid{159} 

\usepackage{xargs} 
\usepackage[colorinlistoftodos,prependcaption,textsize=tiny]{todonotes}
\usepackage{marginnote}

\newcommandx{\todoi}[2][1=]{\todo[inline]{#2}\xspace}
\newcommandx{\todor}[2][1=]{\todo[linecolor=red,backgroundcolor=red!25,bordercolor=red,#1]{sr: #2}\xspace}
\newcommandx{\todog}[2][1=]{\todo[linecolor=red,backgroundcolor=red!25,bordercolor=red,#1]{SG: #2}\xspace}
\newcommandx{\todok}[2][1=]{\todo[linecolor=blue,backgroundcolor=blue!10,bordercolor=blue,#1]{FK: #2}\xspace}
\newcommandx{\todol}[2][1=]{\todo[linecolor=cyan,backgroundcolor=cyan!25,bordercolor=cyan,#1]{ML: #2}\xspace}

\newcommand \ignore[1]{}

\newcommand{\figref}[2][]{Figure#1~\ref{#2}\xspace}
\newcommand{\tabref}[2][]{Table#1~\ref{#2}\xspace}
\newcommand{\eqnref}[2][]{Equation#1~\ref{#2}\xspace}

\newcommand{\dataset}[1]{#1\xspace}

\newcommand{\coco}{\dataset{COCO}}
\newcommand{\tuhoi}{\dataset{TUHOI}}

\newcommand{\ppmi}{\dataset{PPMI}}
\newcommand{\hico}{\dataset{HICO}}
\newcommand{\gold}{\dataset{GOLD}}
\newcommand{\pred}{\dataset{PRED}}

\newcommand{\flickr}{\dataset{Flickr30k}}
\newcommand{\ourdataset}{\dataset{VerSe}}

\newcommand{\argmax}{\operatornamewithlimits{arg\,max}}

\newcommand{\minipagerow}[4]{
\begin{minipage}[b]{0.05\linewidth}
\scriptsize #1
\end{minipage}&
\begin{minipage}[b]{0.15\linewidth}
\includegraphics[height=10mm,width=12mm]{#2}
\end{minipage} &
\begin{minipage}[b]{0.55\linewidth}
\scriptsize #3
\end{minipage} &
\begin{minipage}[b]{0.18\linewidth}
\scriptsize #4
\end{minipage}
}

\newenvironment{checklist}{%
  \begin{list}{}{}
  \let\olditem\item
  \renewcommand\item{\olditem[$\Box$] }
}{%
  \end{list}
}

\title{Unsupervised Visual Sense Disambiguation for Verbs using
  Multimodal Embeddings}

\author{
Spandana Gella, Mirella Lapata and Frank Keller\\
Institute for Language, Cognition and Computation \\
School of Informatics, University of Edinburgh \\
10 Crichton Street, Edinburgh EH8 9AB \\
S.Gella@sms.ed.ac.uk, mlap@inf.ed.ac.uk, keller@inf.ed.ac.uk
}

\date{}

\begin{document}

\maketitle

\begin{abstract}

  We introduce a new task, visual sense disambiguation for verbs:
  given an image and a verb, assign the correct sense of the verb,
  i.e., the one that describes the action depicted in the image. Just
  as textual word sense disambiguation is useful for a wide range of
  NLP tasks, visual sense disambiguation can be useful for multimodal
  tasks such as image retrieval, image description, and text
  illustration. We introduce \ourdataset, a new dataset that augments
  existing multimodal datasets (\coco and \tuhoi) with sense
  labels. We propose an unsupervised algorithm based on Lesk which
  performs visual sense disambiguation using textual, visual, or
  multimodal embeddings. We find that textual embeddings perform well
  when gold-standard textual annotations (object labels and image
  descriptions) are available, while multimodal embeddings perform
  well on unannotated images. We also verify our findings by 
  using the textual and multimodal embeddings as features in a 
  supervised setting and analyse the performance of 
  visual sense disambiguation task. \ourdataset is made 
  publicly available and can be downloaded at: 
  \url{https://github.com/spandanagella/verse}.

\end{abstract}

\section{Introduction}
\label{sec:intro}

\begin{figure}[tb]
\centering
\includegraphics[width=0.3\columnwidth,height=20mm]{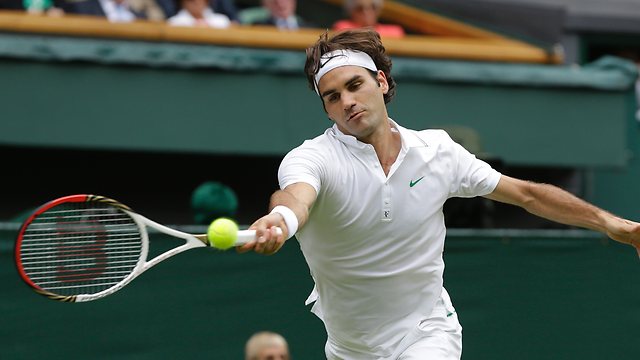}
\includegraphics[width=0.3\columnwidth,height=20mm]{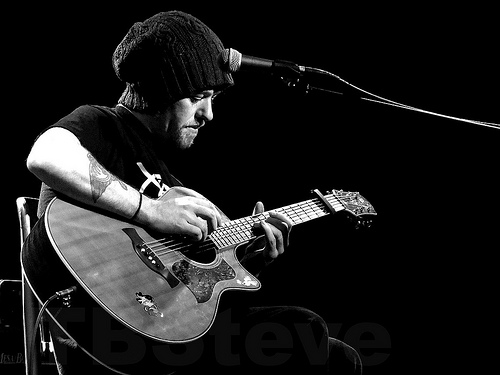}
\includegraphics[width=0.3\columnwidth,height=20mm]{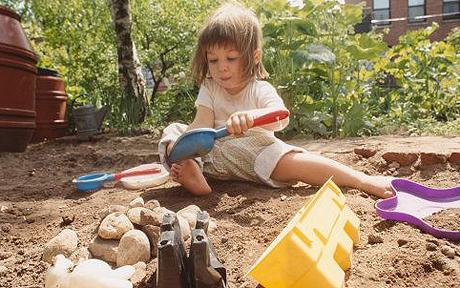}
\caption{Visual sense ambiguity: three of the senses of the verb
  \textit{play}.}
\label{fig:image-sense-examples}
\end{figure}

Word sense disambiguation (WSD) is a widely studied task in natural
language processing: given a word and its context, assign the correct
sense of the word based on a pre-defined sense inventory
\cite{senseval:kilgariff:1998}. WSD is useful for a range of NLP
tasks, including information retrieval, information extraction,
machine translation, content analysis, and lexicography (see
\newcite{Roberto:2009} for an overview). Standard WSD disambiguates
words based on their \emph{textual context}; however, in a multimodal
setting (e.g.,~newspaper articles with photographs), \emph{visual
  context} is also available and can be used for disambiguation. Based
on this observation, we introduce a new task, \emph{visual sense
  disambiguation} (VSD) for verbs: given an image and a verb, assign
the correct sense of the verb, i.e., the one depicted in the
image. While VSD approaches for nouns exist, VSD for verbs is a novel,
more challenging task, and related in interesting ways to \emph{action
  recognition} in computer vision. As an example consider the verb
\textit{play}, which can have the senses \textit{participate in
  sport}, \textit{play on an instrument}, and \textit{be engaged in
  playful activity}, depending on its visual context, see
\figref{fig:image-sense-examples}.

\begin{figure}[tb]
\includegraphics[width=\columnwidth]{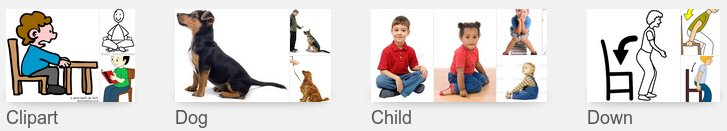}%
\caption{\label{fig:google:sit} Google Image Search trying to
  disambiguate \textit{sit}. All clusters pertain to the \textit{sit
    down} sense, other senses (\textit{baby sit}, \textit{convene})
  are not included.}
\end{figure}

We expect visual sense disambiguation to be useful for multimodal
tasks such as image retrieval. As an example consider the output of
Google Image Search for the query \textit{sit}: it recognizes that the
verb has multiple senses and tries to cluster relevant
images. However, the result does not capture the polysemy of the verb
well, and would clearly benefit from VSD (see
Figure~\ref{fig:google:sit}).

Visual sense disambiguation has previously been attempted for nouns
(e.g.,~\textit{apple} can mean \textit{fruit} or \textit{computer}),
which is a substantially easier task that can be solved with the help
of an object detector
\cite{Barnard:isd:2003,loeff:isd:acl:2006,saenko:isd:nips:2009,chen:isd:2015}.
VSD for nouns is helped by resources such as ImageNet
\cite{imagenet:2009}, a large image database containing 1.4 million
images for 21,841 noun synsets and organized according to the WordNet
hierarchy. However, we are not aware of any previous work on VSD for
verbs, and no ImageNet for verbs exists. Not only image retrieval
would benefit from VSD for verbs, but also other multimodal tasks that
have recently received a lot of interest, such as automatic image
description and visual question answering
\cite{karpathy:2014,msr:caption:generation:2015,vqa:2015}.

In this work, we explore the new task of visual sense disambiguation
for verbs: given an image and a verb, assign the correct sense of the
verb, i.e., the one that describes the action depicted in the image.
We present \ourdataset, a new dataset that augments existing
multimodal datasets (\coco and \tuhoi) with sense labels. \ourdataset
contains 3518 images, each annotated with one of 90~verbs, and the
OntoNotes sense realized in the image. We propose an algorithm based
on the Lesk WSD algorithm in order to perform unsupervised visual
sense disambiguation on our dataset. We focus in particular on how to
best represent word senses for visual disambiguation, and explore the
use of textual, visual, and multimodal embeddings.  Textual embeddings
for a given image can be constructed over object labels or image
descriptions, which are available as gold-standard in the \coco and
\tuhoi datasets, or can be computed automatically using object
detectors and image description models.

Our results show that textual embeddings perform best when
gold-standard textual annotations are available, while multimodal
embeddings perform best when automatically generated object labels are
used. Interestingly, we find that automatically generated image
descriptions result in inferior performance.


\section{Related Work}
\label{sec:related}

There is an extensive literature on word sense disambiguation for
nouns, verbs, adjectives and adverbs. Most of these approaches rely on
lexical databases or sense inventories such as WordNet
\cite{WordNet:Miller:1990} or OntoNotes
\cite{ontonotes:2006}. Unsupervised WSD approaches often rely on
distributional representations, computed over the target word and its
context \cite{lin:1997,diana:2004,brody:2008}.  Most supervised
approaches use sense annotated corpora to extract linguistic features
of the target word (context words, POS tags, collocation features), which
are then fed into a classifier to disambiguate test data
\cite{ims:2010}. Recently, features based on sense-specific semantic
vectors learned using large corpora and a sense inventory such as
WordNet have been shown to achieve state-of-the-art results for
supervised WSD \cite{autoextend:2015,senseretrofit:2015}.

As mentioned in the introduction, all existing work on visual sense
disambiguation has used nouns, starting with
\newcite{Barnard:isd:2003}. Sense discrimination for web images was
introduced by \newcite{loeff:isd:acl:2006}, who used spectral
clustering over multimodal features from the images and web
text. \newcite{saenko:isd:nips:2009} used sense definitions in a
dictionary to learn a latent LDA space overs senses, which they then
used to construct sense-specific classifiers by exploiting the text
surrounding an image.

\begin{table}[t]
\small
\centering
\resizebox{\columnwidth}{!}{
\begin{tabular}{l@{}r@{~}r@{~}r@{~}r@{~}r}
\hline
Dataset & Verbs & Acts & Images & Sen & Des \\ 
\hline
\ppmi \cite{ppmi:2010}  & 2 & 24 & 4800 & N & N\\ 
Stanford 40 Actions \cite{stanford40:2011}  & 33 & 40 & 9532 & N & N\\ 
PASCAL 2012 \cite{pascal:2015} & 9 & 11 & 4588 & N & N \\ 
89 Actions \cite{89actions:2013} & 36 & 89 & 2038 & N & N\\ 
\tuhoi \cite{tuhoi:2014}  & -- & 2974 & 10805 & N & N\\ 
COCO-a \cite{cocoa:2015} & 140 & 162 & 10000 & N & Y\\
\hico \cite{hico:2015} &  111 & 600 & 47774 & Y & N\\ 
\textbf{\ourdataset} (our dataset) & 90 & 163 & 3518 & Y  & Y\\ 
\hline
\end{tabular}}
\caption{Comparison of \ourdataset with existing action recognition datasets.  
  Acts (actions) are verb-object pairs; Sen indicates whether sense
  ambiguity is explicitly handled; Des indicates whether image
  descriptions are included.}
\label{tab:action-dataset-stats}
\end{table}

\subsection{Related Datasets}
\label{sec:datasets}

Most of the datasets relevant for verb sense disambiguation were
created by the computer vision community for the task of human action
recognition (see \tabref{tab:action-dataset-stats} for an overview).
These datasets are annotated with a limited number of actions, where
an action is conceptualized as verb-object pair: \textit{ride horse},
\textit{ride bicycle}, \textit{play tennis}, \textit{play guitar},
etc. Verb sense ambiguity is ignored in almost all action recognition
datasets, which misses important generalizations: for instance, the
actions \textit{ride horse} and \textit{ride bicycle} represent the
same sense of \textit{ride} and thus share visual, textual, and
conceptual features, while this is not the case for \textit{play
  tennis} and \textit{play guitar}. This is the issue we address by
creating a dataset with explicit sense labels.

\ourdataset is built on top of two existing datasets, \tuhoi and
\coco. The Trento Universal Human-Object Interaction (\tuhoi) dataset
contains 10,805 images covering 2974 actions. Action (human-object
interaction) categories were annotated using crowdsourcing: each image
was labeled by multiple annotators with a description in the form of a
verb or a verb-object pair. The main drawback of \tuhoi is that 1576
out of 2974 action categories occur only once, limiting its usefulness
for VSD. The Microsoft Common Objects in Context (\coco) dataset is
very popular in the language/vision community, as it consists of over
120k images with extensive annotation, including labels for 91 object
categories and five descriptions per image. \coco contains no explicit
action annotation, but verbs and verb phrases can be extracted from
the descriptions. 
(But note that not all the \coco images depict actions.)

The recently created Humans Interacting with Common Objects (\hico)
dataset is conceptually similar to \ourdataset. It consists of 47774
images annotated with 111 verbs and 600 human-object interaction
categories. Unlike other existing datasets, \hico uses sense-based
distinctions: actions are denoted by sense-object pairs, rather than
by verb-object pairs. \hico doesn't aim for complete coverage, but
restricts itself to the top three WordNet senses of a verb. The
dataset would be suitable for performing visual sense disambiguation,
but has so far not been used in this way.

\begin{figure}[tb]
\begin{minipage}[b]{\columnwidth}
\scriptsize
\textbf{Verb:} touch
\begin{checklist}
\itemsep-0.6em
\item make physical contact with, possibly with the effect of physically
manipulating. \textcolor{blue}{They touched their fingertips together and
smiled}
\item affect someone emotionally \textcolor{blue}{The president's speech touched
a chord with voters.}
\item be or come in contact without control \textcolor{blue}{They sat so close
that their arms touched.}
\item make reference to, involve oneself with \textcolor{blue}{They had
wide-ranging discussions that touched on the situation in the Balkans.}
\item Achieve a value or quality \textcolor{blue}{Nothing can touch cotton for
durability.}
\item Tinge; repair or improve the appearance of \textcolor{blue}{He touched on
the paintings, trying to get the colors right.}
\end{checklist}
\caption{Example item for depictability and sense annotation: synset
  definitions and examples (in blue) for the verb \textit{touch}.}
\label{fig:visual-sense-annotation-set-up}
\end{minipage}
\end{figure}

\section{\ourdataset Dataset and Annotation}
\label{sec:visual-sense-annotation}

We want to build an unsupervised visual sense disambiguation system,
i.e., a system that takes an image and a verb and returns the correct
sense of the verb. As discussed in Section~\ref{sec:datasets}, most
existing datasets are not suitable for this task, as they do not
include word sense annotation. We therefore develop our own dataset
with gold-standard sense annotation. The Verb Sense (\ourdataset)
dataset is based on \coco and \tuhoi and covers 90 verbs and around
3500 images. \ourdataset serves two main purposes: (1)~to show the
feasibility of annotating images with verb senses (rather than verbs
or actions); (2)~to function as test bed for evaluating automatic
visual sense disambiguation methods.

\paragraph{Verb Selection} Action recognition datasets often use a
limited number of verbs (see Table~\ref{tab:action-dataset-stats}). We
addressed this issue by using images that come with descriptions,
which in the case of action images typically contain verbs. The \coco
dataset includes images in the form of sentences, the \tuhoi dataset
is annotated with verbs or prepositional verb phrases for a given
object
(e.g.,~\textit{sit on} chair), which we use in lieu of descriptions.
We extracted all verbs from all the descriptions in the two datasets
and then selected those verbs that have more than one sense in the
OntoNotes dictionary, which resulted in 148 verbs in total (94 from
\coco and 133 from \tuhoi).

\paragraph{Depictability Annotation} A verb can have multiple senses,
but not all of them may be depictable, e.g., senses describing
cognitive and perception processes. Consider two senses of
\emph{touch}: \textit{make physical contact} is depictable, whereas
\textit{affect emotionally} describes a cognitive process and is not
depictable. We therefore need to annotate the synsets of a verb as
depictable or non-depictable. Amazon Mechanical Turk (AMT) workers
were presented with the definitions of all the synsets of a verb,
along with examples, as given by OntoNotes. An example for this
annotation is shown in \figref{fig:visual-sense-annotation-set-up}. We
used OntoNotes instead of WordNet, as WordNet senses are very
fine-grained and potentially make depictability and 
sense annotation (see below) harder. Granularity issues with WordNet
for text-based WSD are well documented \cite{Roberto:2009}.


OntoNotes lists a total of 921 senses for our 148 target verbs. For
each synset, three AMT workers selected all depictable senses. The
majority label was used as the gold standard for subsequent
experiments. This resulted in a 504 depictable senses. Inter-annotator
agreement (ITA) as measured by Fleiss' Kappa was $0.645$.

\begin{table}[t]
\tabcolsep 3pt
\resizebox{\columnwidth}{!}{
\begin{tabular}{l@{~}l@{~}r@{~}r@{~}r@{~}r@{~}r}
\hline
Verb type & Examples & Verbs & Images & Senses & Depct & ITA\\ 
\hline
Motion & run, walk, jump, etc. & 39 & 1812 & 10.76 & 5.79 & 0.680\\ 
Non-motion & sit, stand, lay, etc. & 51 & 1698 & 8.27 & 4.86& 0.636 \\ 
\hline
\end{tabular}
}
\caption{Overview of \ourdataset dataset divided 
  into motion and non-motion verbs; Depct: depictable senses; ITA:
  inter-annotator agreement.}
\label{tab:data-stats}
\end{table}

\paragraph{Sense Annotation} We then annotated a subset of the images
in \coco and \tuhoi with verb senses. For every image we assigned the
verb that occurs most frequently in the descriptions for that image
(for \tuhoi, the descriptions are verb-object pairs, see above).
However, many verbs are represented by only a few images, while a few
verbs are represented by a large number of images. The datasets
therefore show a Zipfian distribution of linguistic units, which is
expected and has been observed previously for \coco
\cite{cocoa:2015}. For sense annotation, we selected only verbs for
which either \coco or \tuhoi contained five or more images, resulting
in a set of 90 verbs (out of the total~148).  All images for these
verbs were included, giving us a dataset of 3518 images: 2340 images
for 82 verbs from \coco and 1188 images for 61 verbs from \tuhoi (some
verbs occur in both datasets).

These image-verb pairs formed the basis for sense annotation. AMT
workers were presented with the image and all the depictable OntoNotes
senses of the associated verb. The workers had to chose the sense of
the verb that was instantiated in the image (or ``none of the above'',
in the case of irrelevant images). Annotators were given sense
definitions and examples, as for the depictability annotation (see
Figure~\ref{fig:visual-sense-annotation-set-up}). For every image-verb
pair, five annotators performed the sense annotation task.  A total of
157 annotators participated, reaching an inter-annotator agreement of
$0.659$ (Fleiss' Kappa). Out of 3528 images, we discarded 18 images
annotated with ``none of the above'', resulting in a set of 3510
images covering 90 verbs and 163 senses. We present statistics of our
dataset in \tabref{tab:data-stats}; we group the verbs into motion
verbs and non-motion verb using \newcite{levin:1993} classes.





\section{Visual Sense Disambiguation}
\label{sec:method}

For our disambiguation task, we assume we have a set of images $I$,
and a set of polysemous verbs $V$ and each image $i \in I$ is paired
with a verb $v \in V$. For example, \figref{fig:image-sense-examples}
shows different images paired with the verb \emph{play}. Every verb $v
\in V$, has a set of senses $\mathcal{S}(v)$, described in a
dictionary $\mathcal{D}$. Now given an image $i$ paired with a verb
$v$, our task is to predict the correct sense $\hat{s} \in
\mathcal{S}(v)$, i.e.,~the sense that is depicted by the associated
image. Formulated as a scoring task, disambiguation consists of
finding the maximum over a suitable scoring function~$\Phi$:
%
\begin{equation} \label{eq:generic}
\hat{s} = \argmax_{s \in \mathcal{S}(v)} \Phi(s,i,v,\mathcal{D})
\end{equation}
%
For example, in \figref{fig:image-sense-examples}, the correct sense
for the first image is \textit{participate in sport}, for the second
one it is \textit{play on an instrument},~etc.

The \newcite{lesk:1986} algorithm is a well known knowledge-based
approach to WSD which relies on the calculation of the word overlap
between the sense definition and the context in which a word
occurs. It is therefore an unsupervised approach, i.e., it does not
require sense-annotated training data, but instead exploits resources
such as dictionaries or ontologies to infer the sense of a word in
context. Lesk uses the following scoring function to disambiguate the
sense of a verb~$v$:
\begin{equation} \label{eq:lesk-original}
\begin{split}
  \Phi(s,v,\mathcal{D}) = |\textit{context}(v) \cap
  \textit{definition}(s,\mathcal{D})|
\end{split}
\end{equation}
Here, $\mathit{context}(v)$ the set of words that occur close the
target word $v$ and $\textit{definition}(s,\mathcal{D})$ is the set of
words in the definition of sense $s$ in the dictionary~$\mathcal{D}$.
Lesk's approach is very sensitive to the exact wording of definitions
and results are known to change dramatically for different sets of
definitions \cite{Roberto:2009}. Also, sense definitions are often
very short and do not provide sufficient vocabulary or context.

\begin{figure}[tb]
\includegraphics[width=\columnwidth]{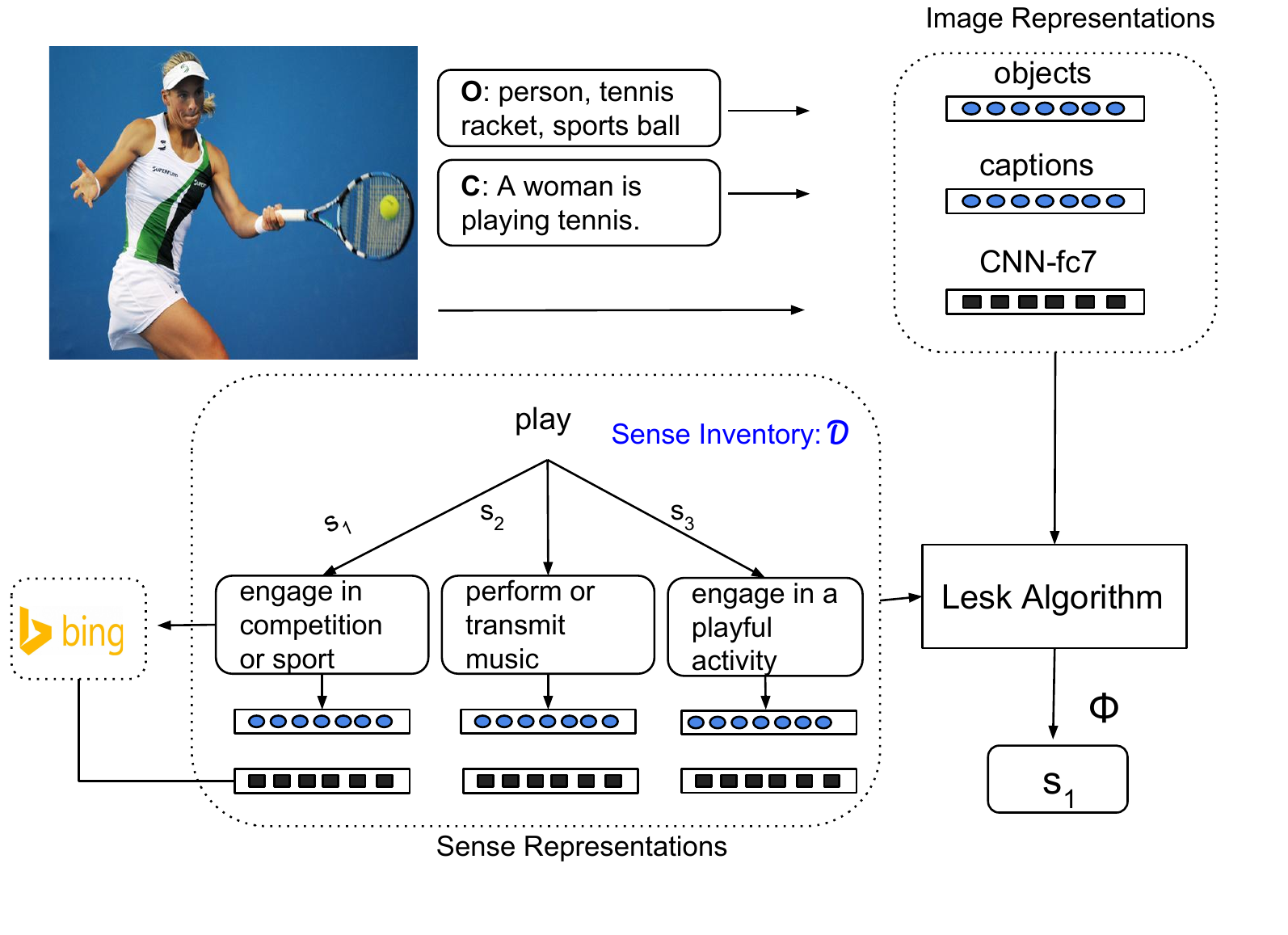}%
\caption{Schematic overview of the visual sense disambiguation model.}
\label{fig:methodology} 
\end{figure}

We propose a new variant of the Lesk algorithm to disambiguate the
verb sense that is depicted in an image. In particular, we explore the
effectiveness of textual, visual and multimodal representations in
conjunction with Lesk.  An overview of our methodology is given in
\figref{fig:methodology}. For a given image $i$ labeled with verb $v$
(here \textit{play}), we create a representation (the
vector~$\mathbf{i}$), which can be text-based (using the object labels
and descriptions for~$i$), visual, or multimodal. Similarly, we create
text-based, visual, and multimodal representations (the
vector~$\mathbf{s}$) for every sense $s$ of a verb. Based on the
representations $\mathbf{i}$ and $\mathbf{s}$ (detailed below), we can
then score senses as:\footnote{Taking the dot product of two
  normalized vectors is equivalent to using cosine as similarity
  measure. We experimented with other similarity measures, but cosine
  performed best.}
\begin{equation} \label{eq:lesk-scoring-function}
\Phi(s,v,i,\mathcal{D}) = \mathbf{i} \cdot \mathbf{s}
\end{equation}
Note that this approach is unsupervised: it requires no sense
annotated training data; we will use the sense annotations in our
\ourdataset dataset only for evaluation.

\begin{figure}[tb]
    \vspace{-3ex}
    \centering
	\includegraphics[width=\linewidth]{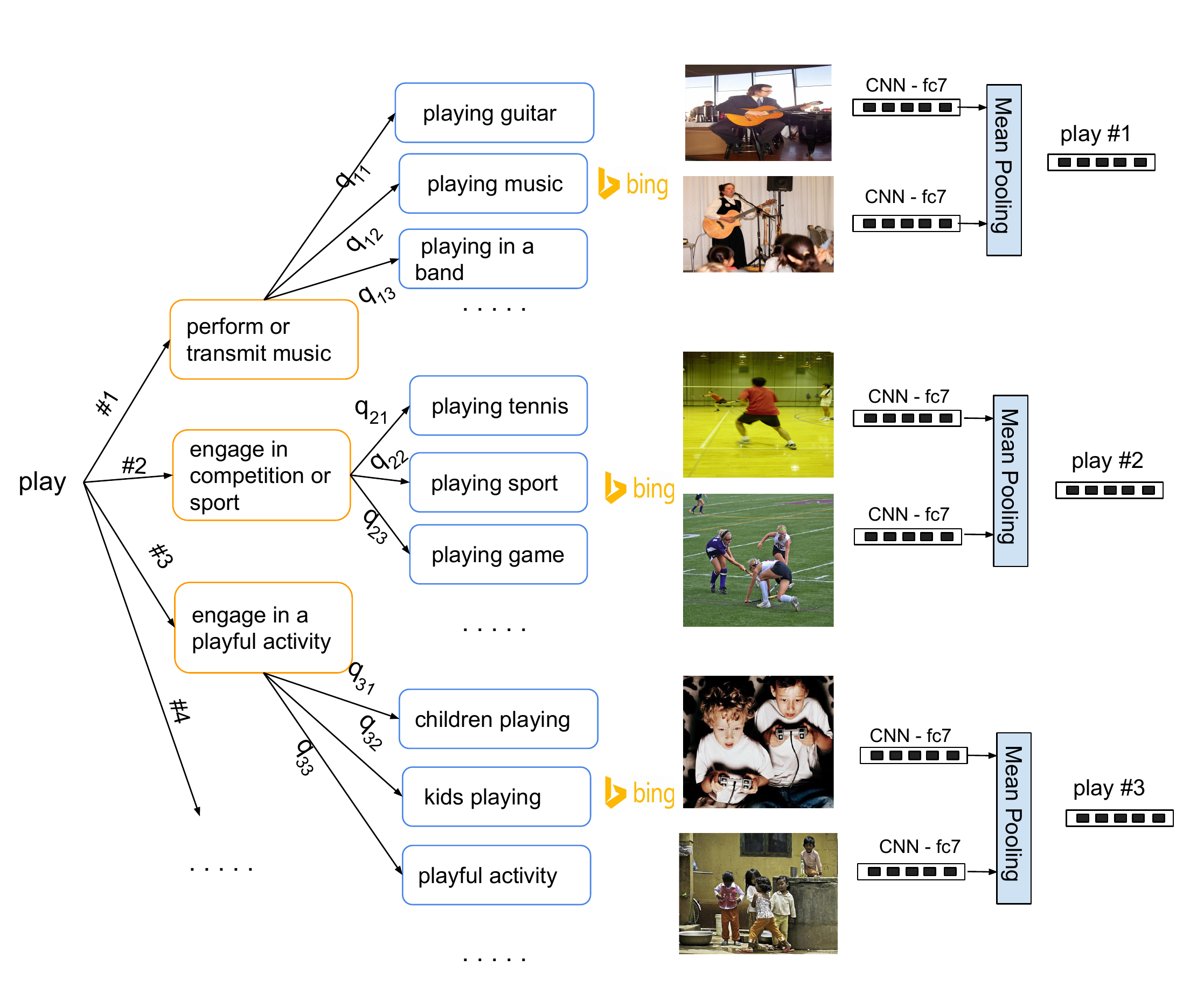}       
	\caption{Extracting visual sense representation for the verb \textit{play}.}
	\label{fig:sense-cnn-rep}
\end{figure}

\subsection{Sense Representations}
\label{ssec:sense-rep}

For each candidate verb sense, we create a text-based sense
representation~$\mathbf{s^t}$ and a visual sense
representation~$\mathbf{s^c}$.

\paragraph{Text-based Sense Representation} We create a vector
$\mathbf{s^t}$ for every sense $s \in \mathcal{S}(v)$ of a verb $v$
from its definition and the example usages provided in the OntoNotes
dictionary $\mathcal{D}$. We apply word2vec \cite{word2vec:2013}, a
widely used model of word embeddings, to obtain a vector for every
content word in the definition and examples of the sense. We then take
the average of these vectors to compute an overall representation of
the verb sense. For our experiments we used the pre-trained
300~dimensional vectors available with the word2vec package (trained
on part of Google News dataset, about 100 billion words).

\paragraph{Visual Sense Representation} Sense dictionaries typically
provide sense definitions and example sentences, but no visual
examples or images. For nouns, this is remedied by ImageNet
\cite{imagenet:2009}, which provides a large number of example images
for a subset of the senses in the WordNet noun hierarchy. However, no
comparable resource is available for verbs (see
Section~\ref{sec:datasets}).

In order to obtain visual sense representation $\mathbf{s^c}$, we
therefore collected sense-specific images for the verbs in our
dataset.  For each verb sense $s$, three trained annotators were
presented with the definition and examples from OntoNotes, and had to
formulate a query $\mathcal{Q}(s)$ that would retrieve images
depicting the verb sense when submitted to a search engine.  For every
query $q$ we retrieved images $\mathcal{I}(q)$ using Bing image search
(for examples, see \figref{fig:sense-cnn-rep}).  We used the top 50
images returned by Bing for every query.


Once we have images for every sense, we can turn these images into
feature representations using a convolutional neural network (CNN).
Specifically, we used the VGG 16-layer architecture (VGGNet) trained
on 1.2M images of the 1000 class ILSVRC~2012 object classification
dataset, a subset of ImageNet \cite{vgg:2014}. This CNN model has a
top-5 classification error of 7.4\% on ILSVRC~2012. We use the
publicly available reference model implemented using CAFFE
\cite{caffe:2014} to extract the output of the fc7 layer, i.e., a 4096
dimensional vector $\mathbf{c_i}$, for every image $i$. We perform
mean pooling over all the images extracted using all the queries of a
sense to generate a single visual sense representation $\mathbf{s^c}$
(shown in \eqnref{eq:vis-sense-rep}):
\begin{equation} \label{eq:vis-sense-rep}
\mathbf{s^c} = \frac{1}{n} \sum_{q_j \in \mathcal{Q}(s)} \sum_{i \in
\mathcal{I}(q_j)} \mathbf{c_i}
\end{equation}
where $n$ is the total number of images retrieved per sense~$s$.

%

\subsection{Image Representations}
\label{ssec:feature-rep}


We first explore the possibility of representing the image indirectly,
viz., through text associated with it in the form of object labels or
image descriptions (as shown in \figref{fig:methodology}). We
experiment with two different forms of textual annotation: \gold
annotation, where object labels and descriptions are provided by human
annotators, and predicted (\pred) annotation, where state-of-the-art
object recognition and image description generation systems are
applied to the image.
%
\paragraph{\textbf{Object Labels (O)}} \gold object annotations are provided
  with the two datasets we use.  Each image sampled from \coco is
  annotated with one or more of 91 object categories.  Each image from
  \tuhoi is annotated with one more of 189 object categories. \pred
  object annotations were generated using the same VGG-16-layer CNN
  object recognition model that was used to compute visual sense
  representations. Only object labels with object detection threshold
  of $t>0.2$ were used.
\paragraph{\textbf{Descriptions (C)}} To obtain \gold image descriptions, we
  used the used human-generated descriptions that come with \coco. For
  \tuhoi images, we generated descriptions of the form
  subject-verb-object, where the subject is always \textit{person},
  and the verb-object pairs are the action labels that come with
  \tuhoi.  To obtain \pred descriptions, we generated three
  descriptions for every image using the state-of-the-art image
  description system of \newcite{google:show-tell:2014}.\footnote{We
    used Karpathy's implementation, publicly available at
    \url{https://github.com/karpathy/neuraltalk}.}
%

%
We can now create a textual representation $\mathbf{i^t}$ of the
image~$i$. Again, we used word2vec to obtain word embeddings, but
applied these to the object labels and to the words in the image
descriptions. An overall representation of the image is then computed
by averaging these vectors over all labels, all content words in the
description, or both.

Creating a visual representation $\mathbf{i^c}$ of an image~$i$ is
straightforward: we extract the fc7 layer of the VGG-16 network when
applied to the image and use the resulting vector as our image
representation (same setup as in Section~\ref{ssec:sense-rep}).

Apart from experimenting with separate textual and visual
representations of images, it also makes sense to combine the two
modalities into a multimodal representation. The simplest approach is
a concatenation model which appends textual and visual features. More
complex multimodal vectors can be created using methods such as
Canonical Correlation Analysis (CCA) and Deep Canonical Correlation
Analysis (DCCA) \cite{cca:2004,dcca:2013,dcca:2015}.  CCA allows us to
find a latent space in which the linear projections of text and image
vectors are maximally correlated
\cite{transfer:cca:2014,hodosh:2015}. DCCA can be seen as non-linear
version of CCA and has been successfully applied to image description task 
\cite{dcca:image-text:mapping:2015}, outperforming previous
approaches, including kernel-based CCA.

We use both CCA and DCCA to map the vectors $\mathbf{i^t}$ and
$\mathbf{i^c}$ (which have different dimensions) into a joint latent
space of $n$ dimensions. We represent the projected vectors of textual
and visual features for image $i$ as $\mathbf{{i^t}^\prime}$ and
$\mathbf{{i^c}^\prime}$ and combine them to obtain multimodal
representation $\mathbf{i^m}$ as follows:
\begin{equation} \label{eq:weighted-average-image}
\mathbf{i^m} = \lambda_t \mathbf{{i^t}^\prime} + \lambda_c \mathbf{{i^c}^\prime}
\end{equation}
We experimented with a number of parameter settings for $\lambda_t$
and $\lambda_c$ for textual and visual models respectively. We use the
same model to combine the multimodal representation for sense $s$ as
follows:
\begin{equation} \label{eq:weighted-average-sense}
\mathbf{s^m} = \lambda_t \mathbf{{s^t}^\prime} + \lambda_c \mathbf{{s^c}^\prime}
\end{equation}

We use these vectors ($\mathbf{i^t}$, $\mathbf{s^t}$), 
($\mathbf{i^c}$, $\mathbf{s^c}$) and ($\mathbf{i^m}$, $\mathbf{s^m}$) 
as described in  ~\eqnref{eq:lesk-scoring-function} to perform 
sense disambiguation.

\begin{table*}[!htb]
\small
\tabcolsep 5pt
\resizebox{\textwidth}{!}{

\begin{tabular}{lccccccccccccccccc}
\multicolumn{18}{c}{(a) Motion verbs (39), FS: 70.8,  MFS: 86.2} \\
\hline
Annotation &  \multicolumn{3}{c}{Textual} & & Vis & &
\multicolumn{3}{c}{Concat \scriptsize{(CNN+)}} & & \multicolumn{3}{c}{CCA
\scriptsize{(CNN+)}} & & \multicolumn{3}{c}{DCCA \scriptsize{(CNN+)}}\\ 
 \cline{2-4}
 \cline{6-6}
 \cline{8-10}
 \cline{12-14}
 \cline{16-18}
& \scriptsize{O}& \scriptsize{C}& \scriptsize{O+C}& & \scriptsize{CNN} & &
\scriptsize{O} & \scriptsize{C} & \scriptsize{O+C} & & \scriptsize{O} &
\scriptsize{C} & \scriptsize{O+C}  &  & \scriptsize{O} & \scriptsize{C} &
\scriptsize{O+C} \\
\hline
\gold &  54.6 & \textbf{73.3} & \textbf{75.6} & & 58.3 & & 66.6 & \textbf{74.7}
& \textbf{73.8} & & 50.5 & \textbf{75.4} & \textbf{74.0} & & 52.4 & 66.3 &
68.3\\ 
\pred & 65.1 & 54.9 & 61.6 & & 58.3 & & \textbf{72.6} & 63.6 & 66.5 & &  54.0&
56.6 & 56.2 & &  57.1 & 56.5 & 56.2 \\ 

\hline

& \\ 

\multicolumn{18}{c}{(b) Non-motion verbs (51), FS: 80.6, MFS: 90.7} \\

\hline
Annotation &  \multicolumn{3}{c}{Textual} & & Vis & &
\multicolumn{3}{c}{Concat \scriptsize{(CNN+)}} & & \multicolumn{3}{c}{CCA
\scriptsize{(CNN+)}} & & \multicolumn{3}{c}{DCCA \scriptsize{(CNN+)}}\\ 
 \cline{2-4}
 \cline{6-6}
 \cline{8-10}
 \cline{12-14}
 \cline{16-18}
& \scriptsize{O}& \scriptsize{C}& \scriptsize{O+C}& & \scriptsize{CNN} & &
\scriptsize{O} & \scriptsize{C} & \scriptsize{O+C} & & \scriptsize{O} &
\scriptsize{C} & \scriptsize{O+C}  &  & \scriptsize{O} & \scriptsize{C} &
\scriptsize{O+C} \\
\hline
\gold &  57.0 & 72.7 & 72.6 & & 56.1 & & 66.0 & 72.2
& 71.3 & & 53.6 & 71.6 & 70.2 & & 57.3 & 59.8 & 55.1\\ 
\pred & 59.0 & 64.3 & 64.0 & & 56.1 & & 63.8 & 66.3 & 66.1 & &  50.7 & 
55.3 & 54.8 & &  49.5 & 50.0 & 50.0 \\  

\hline

\end{tabular}}
\caption{Accuracy scores for motion and non-motion verbs using for
  different types of sense and image representations (O: object
  labels, C: image descriptions, CNN: image features, FS: first sense
  heuristic, MFS: most frequent sense heuristic). Configurations that
  performed better than FS in \textbf{bold}.}
\label{tab:scores-verbdataset}
\end{table*}

\section{Experiments}
\label{sec:experiments}

\subsection{Unsupervised Setup}
\label{sec:setup}

To train the CCA and DCCA models, we use the text representations
learned from image descriptions of \coco and \flickr dataset as one
view and the VGG-16 features from the respective images as the second
view. We divide the data into train, test and development samples
(using a 80/10/10 split). We observed that the correlation scores for
DCCA model were better than for the CCA model. We use the trained
models to generate the projected representations of text and visual
features for the images in \ourdataset. Once the textual and visual
features are projected, we then merge them to get the multimodal
representation. We experimented with different ways of combining
visual and textual features projected using CCA or DCCA: (1)~weighted
interpolation of textual and visual features (see
Equations~\ref{eq:weighted-average-image} and
\ref{eq:weighted-average-sense}), and (2)~concatenating the vectors of
textual and visual features.

To evaluate our proposed method, we compare against the first sense
heuristic, which defaults to the sense listed first in the dictionary
(where senses are typically ordered by frequency). This is a strong
baseline which is known to outperform more complex models in
traditional text-based WSD.  In \ourdataset we observe skewness in the
distribution of the senses and the first sense heuristic is as strong
as over text. Also the most frequent sense heuristic, which assigns
the most frequently annotated sense for a given verb in \ourdataset,
shows very strong performance. It is supervised (as it requires sense
annotated data to obtain the frequencies), so it should be regarded as
an upper limit on the performance of the unsupervised methods we
propose (also, in text-based WSD, the most frequent sense heuristic is
considered an upper limit, \newcite{Roberto:2009}).


\subsubsection{Results}
\label{ssec:results}

In \tabref{tab:scores-verbdataset}, we summarize the results of the
gold-standard (\gold) and predicted (\pred) settings for motion and
non-motion verbs across representations. In the \gold setting we find
that for both types of verbs, textual representations based on image
descriptions (C) outperform visual representations (CNN features). The
text-based results compare favorably to the original Lesk 
(as described in  ~\eqnref{eq:lesk-original}), which performs at 30.7 for motion verbs and 
36.2 for non-motion verbs in the \gold setting. This improvement 
is clearly due to the use of word2vec embeddings.\footnote{We 
also experimented with Glove vectors \cite{glove:2014} but 
observed that word2vec representations consistently achieved 
better results that Glove vectors.} Note that CNN-based 
visual features alone performed better than gold-standard 
object labels alone in the case of motion verbs.

We also observed that adding visual features to textual features
improves performance in some cases: multimodal features perform better
than textual features alone both for object labels (CNN+O) and for
image descriptions (CNN+C). However, adding CNN features to textual
features based on object labels and descriptions together (CNN+O+C)
resulted in a small decrease in performance. Furthermore, we note that
CCA models outperform simple vector concatenation in case of \gold
setting for motion verbs, and overall DCCA performed considerably
worse than concatenation. Note that for CCA and DCCA we report the
best performing scores achieved using weighted interpolation of
textual and visual features with weights $\lambda_t=0.5$ and
$\lambda_c=0.5$.

When comparing to our baseline and upper limit, we find that the all
the \gold models which use descriptions-based representations (except
DCCA) outperform to the first sense heuristic for motion-verbs
(accuracy 70.8), whereas they performed below the first sense heuristic
in case of non-motion verbs (accuracy 80.6). As expected, both motion
and non-motion verbs performed significantly below the most frequent
sense heuristic (accuracy 86.2 and 90.7 respectively), which we argued
provides an upper limit for unsupervised approaches.

\begin{table}[t]
\begin{tabular}{lcccccc}
\hline
\multicolumn{7}{c}{Motion verbs (19), FS: 60.0,  MFS: 76.1 } \\
\hline
Features & & \multicolumn{2}{c}{\gold} & & \multicolumn{2}{c}{\pred} \\
\cline{3-4}
\cline{6-7}
& & \scriptsize{Sup} & \scriptsize{Unsup} & & \scriptsize{Sup} & \scriptsize{Unsup} \\
\hline
O &  & 82.3 & 35.3 &  & 80.0& 43.8 \\
C &  & 78.4 & 53.8 & & 69.2& 41.5 \\
O+C &  & 80.0 & 55.3 &  & 70.7& 45.3 \\
CNN &  & 82.3  & 58.4 &  & 82.3 & 58.4 \\
CNN+O &    & 83.0 & 48.4  & & 83.0 & 60.0 \\
CNN+C &   & 82.3 & 66.9 &  & 82.3 & 53.0 \\
CNN+O+C &  & 83.0 & 58.4 &  & 83.0 & 55.3 \\
\hline

\end{tabular}
\caption{Accuracy scores for motion verbs for both supervised and unsupervised approaches using
  different types of sense and image representation features. 
  }
\label{tab:scores-sup-unsup-motion}
\end{table}

\begin{table}[t]
\begin{tabular}{lcccccc}
\hline
\multicolumn{7}{c}{Non-Motion verbs (19), FS: 71.3,  MFS: 80.0 } \\
\hline
Features & & \multicolumn{2}{c}{\gold} & & \multicolumn{2}{c}{\pred} \\
\cline{3-4}
\cline{6-7}
& & \scriptsize{Sup} & \scriptsize{Unsup} & & \scriptsize{Sup} & \scriptsize{Unsup} \\
\hline
O &  & 79.1 & 48.6 &  & 78.2& 46.0 \\
C &  & 79.1 & 53.9 &  & 77.3 & 61.7 \\
O+C &  & 79.1 & 66.0 &  & 77.3 & 55.6 \\
CNN &  & 80.0  & 55.6 &  & 80.0 & 55.6 \\
CNN+O &    & 80.0 & 56.5  & & 80.0 & 52.1 \\
CNN+C &   & 80.0 & 56.5&  & 80.3 & 60.0 \\
CNN+O+C &  & 80.0 & 59.1 &  & 80.0 & 55.6 \\
\hline

\end{tabular}
\caption{Accuracy scores for non-motion verbs for both supervised and unsupervised approaches using
  different types of sense and image representation features. 
  }
\label{tab:scores-sup-unsup-non-motion}
\end{table}

We now turn the \pred configuration, i.e., to results obtained using
object labels and image descriptions predicted by state-of-the-art
automatic systems. This is arguably the more realistic scenario, as it
only requires images as input, rather than assuming human-generated
object labels and image descriptions (though object detection and
image description systems are required instead).  In the \pred
setting, we find that textual features based on object labels (O)
outperform both first sense heuristic and textual features based on
image descriptions (C) in the case of motion verbs.  Combining textual
and visual features via concatenation improves performance for both
motion and non-motion verbs. The overall best performance of 72.6 for
predicted features is obtained by combining CNN features and
embeddings based on object labels and outperforms first sense
heuristic in case of motion verbs (accuracy 70.8).  In the \pred
setting for both classes of verbs the simpler concatenation model
performed better than the more complex CCA and DCCA models. Note that
for CCA and DCCA we report the best performing scores achieved using
weighted interpolation of textual and visual features with weights
$\lambda_t=0.3$ and $\lambda_c=0.7$. Overall, our findings are
consistent with the intuition that motion verbs are easier to
disambiguate than non-motion verbs, as they are more depictable and
more likely to involve objects. Note that this is also reflected in
the higher inter-annotator agreement for motion verbs (see
\tabref{tab:data-stats}).

\subsection{Supervised Experiments and Results}

Along with the unsupervised experiments we investigated the 
performance of textual and visual representations of 
images in a simplest supervised setting. We trained logistic 
regression classifiers for sense prediction by dividing the 
images in \ourdataset dataset into train and test splits. 
To train the classifiers we selected all the verbs which has atleast 
20 images annotated and has at least two senses in \ourdataset. 
This resulted in 19 motion verbs and 19 non-motion verbs. 
Similar to our unsupervised experiments we explore multimodal 
features by using both textual and visual features for classification 
(similar to concatenation in unsupervised experiments).

In \tabref{tab:scores-sup-unsup-motion} we report accuracy scores for 
19 motion verbs using a supervised logistic regression classifier 
and for comparison we also report the scores of our proposed 
unsupervised algorithm for both \gold and \pred setting. Similarly in 
\tabref{tab:scores-sup-unsup-non-motion} we report the accuracy 
scores for 19 non-motion verbs. We observe that all supervised 
classifiers for both motion and non-motion verbs performing better 
than first sense baseline. Similar to our findings using an 
unsupervised approach we find that in most cases multimodal 
features obtained using concatenating textual and visual features 
has outperformed textual or visual features alone especially 
in the \pred setting which is arguably the more realistic scenario. 
We observe that the features from \pred image descriptions showed 
better results for non-motion verbs for both supervised and 
unsupervised approaches whereas \pred object features showed better 
results for motion verbs. We also observe that supervised classifiers 
outperform most frequent sense for motion verbs and for non-motion 
verbs our scores match with most frequent sense heuristic. 




\begin{table}[t]
\begin{tabular}{@{~}l@{~~}l@{~}l@{~~}l@{~}}
\hline 
{\scriptsize{Verb}} & {\scriptsize{Image}} &
{\scriptsize{Predicted Descriptions}} &{\scriptsize{Pred. Obj.}}\\
\hline
\minipagerow{play}{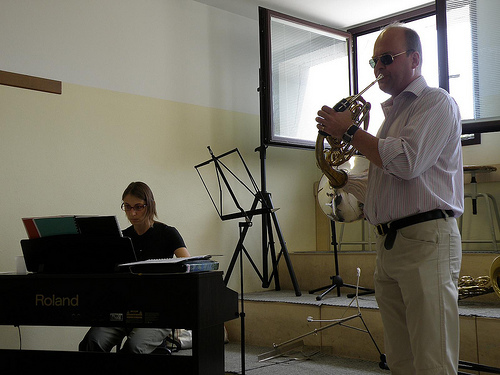}{%
A man holding a nintendo wii game controller. A man and a woman
playing a video game. A man and a woman are playing a video
game.}{person, bassoon, violin fiddle, oboe, hautboy} \\
\hline
\minipagerow{swing}{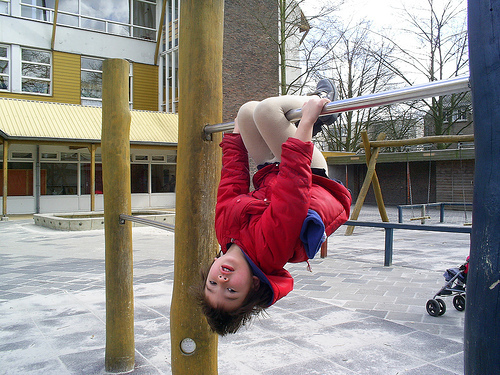}{A woman standing
  next to a fire hydrant.
A woman walking down a street holding an umbrella.
A woman standing on a sidewalk holding an umbrella.}{person,
horizontal bar, high bar, pole} \\
\hline
\minipagerow{feed}{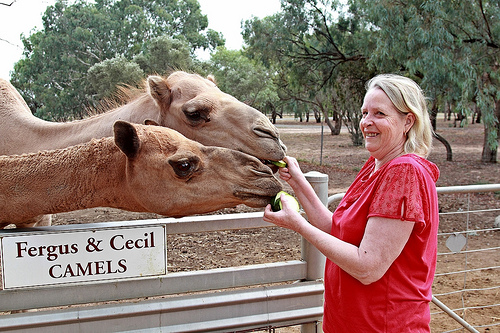}{A couple of cows
  standing next to each other. A cow that is standing in the dirt.
A close up of a horse in a stable}{
arabian camel, dromedary, person} \\
\hline
\end{tabular}
\caption{Images that were assigned an incorrect sense in the \pred setting.}
\label{tab:error-analysis}
\end{table}

\subsection{Error Analysis} 
\label{sec:errors}

In order to understand the cases where the proposed unsupervised algorithm failed,
we analyzed the images that were disambiguated incorrectly. For the
\pred setting, we observed that using predicted image descriptions
yielded lower scores compared to predicted object labels. The main
reason for this is that the image description system often generates
irrelevant descriptions or descriptions not related to the action
depicted, whereas the object labels predicted by the CNN model tend to
be relevant. This highlights that current image description systems
still have clear limitations, despite the high evaluation scores
reported in the literature
\cite{google:show-tell:2014,msr:caption:generation:2015}.  Examples
are shown in \tabref{tab:error-analysis}: in all cases human generated
descriptions and object labels that are relevant for disambiguation, which
explains the higher scores in the \gold setting.

\section{Conclusion}
\label{sec:conclusions}

We have introduced the new task of visual verb sense disambiguation:
given an image and a verb, identify the verb sense depicted in the
image. We developed the new \ourdataset dataset for this task, based
on the existing \coco and \tuhoi datasets. We proposed an unsupervised
visual sense disambiguation model based on the Lesk algorithm and
demonstrated that both textual and visual information associated with
an image can contribute to sense disambiguation. In an in-depth
analysis of various image representations we showed that object labels
and visual features extracted using state-of-the-art convolutional
neural networks result in good disambiguation performance, while
automatically generated image descriptions are less useful. 


\bibliography{references}
\bibliographystyle{naaclhlt2016}

\end{document}